\pdfoutput=1
\documentclass[11pt]{article}

\usepackage[]{acl}

\usepackage{times}
\usepackage{latexsym}
\usepackage{amsfonts}
\usepackage{amsmath}
\usepackage{amssymb} 
\usepackage[ruled,linesnumbered]{algorithm2e}
\usepackage{algpseudocode}
\usepackage{makecell}
\usepackage{booktabs}
\usepackage{multirow}
\usepackage{graphicx}
\usepackage{float}
\usepackage[capitalise]{cleveref}
\usepackage{diagbox}
\usepackage{caption}
\usepackage{subcaption}

\crefname{section}{\S}{\S\S}
\Crefname{section}{\S}{\S\S}    % must define start-of-sentence version explicitly since \S isn't a letter
\crefformat{section}{#2\S#1#3}  % remove space between section symbol and the number, and include \S in the hyperlink
\Crefformat{section}{#2\S#1#3}
\crefrangeformat{section}{\S\S#3#1#4--#5#2#6}
\Crefrangeformat{section}{\S\S#3#1#4--#5#2#6}

\usepackage[T1]{fontenc}
\usepackage[utf8]{inputenc}

\usepackage{microtype}

\newcommand\PET{\textsc{DeT}\xspace}
\newcommand\PETs{\textsc{DeT}s\xspace}

\DeclareMathOperator*{\argmin}{arg\,min}

\usepackage{enumitem}

\title{\textit{Different Tunes Played with Equal Skill:}\\Exploring a Unified Optimization Subspace for Delta Tuning}

\author{
 Jing~Yi$^{1}\thanks{\ \ Indicates equal contribution.}$\hspace{0.5em}, Weize~Chen$^{1*}$, Yujia~Qin$^{1*}$, Yankai~Lin$^{2,3}$, Ning~Ding$^{1}$, Xu~Han$^{1}$, \\ \textbf{Zhiyuan~Liu}$^{1,4,5}\thanks{\ \  Corresponding author.}$\hspace{0.5em}, \textbf{Maosong~Sun$^{1,4,5\dag}$, Jie~Zhou$^6$} \\
 $^1$NLP Group, DCST, IAI, BNRIST, Tsinghua University, Beijing \\
 $^2$Gaoling School of Artificial Intelligence, Renmin University of China, Beijing \\
 $^3$Beijing Key Laboratory of Big Data Management and Analysis Methods, Beijing \\
 $^4$International Innovation Center of Tsinghua University, Shanghai \\
 $^5$Quan Cheng Laboratory $^6$Pattern Recognition Center, WeChat AI, Tencent Inc. \\
\texttt{\{yi-j20, chenwz21, qyj20\}@mails.tsinghua.edu.cn}\\
\texttt{\{liuzy,sms\}@tsinghua.edu.cn}
}

\begin{document}
\maketitle

\begin{abstract}
Delta tuning (\PET, also known as parameter-efficient tuning) is deemed as the new paradigm for using pre-trained language models (PLMs). Up to now, various \PETs with distinct design elements have been proposed, achieving performance on par with fine-tuning. However, the mechanisms behind the above success are still under-explored, especially the connections among various \PETs. To fathom the mystery, we hypothesize that the adaptations of different \PETs could all be reparameterized as low-dimensional optimizations in a unified optimization subspace, which could be found by jointly decomposing independent solutions of different \PETs. Then we explore the connections among different \PETs by conducting optimization within the subspace. In experiments, we find that, for a certain \PET, conducting optimization simply in the subspace could achieve comparable performance to its original space, and the found solution in the subspace could be transferred to another \PET and achieve non-trivial performance. We also visualize the performance landscape of the subspace, and find that, there exists a substantial region where different \PETs all perform well. Finally, we extend our analysis and show the strong connections 
between fine-tuning and \PETs. The codes are publicly available at \url{https://github.com/thunlp/Unified-DeltaTuning}.

\end{abstract}
\section{Introduction}
Serving as the critical backbone for NLP, pre-trained language models (PLMs) achieve superior performance when adapted to downstream tasks~\citep{han2021pre}. Conventionally, the dominant way for such an adaptation is fine-tuning, which requires updating and storing all the parameters in PLMs. Consequently, with ever-larger PLMs continually being proposed~\citep{raffel2020exploring,NEURIPS2020_1457c0d6}, fine-tuning becomes extremely computationally expensive. As an alternative, various delta tuning algorithms (\PETs) spring up, which freeze most of the parameters and only optimize minimal adaptive parameters~\citep{ding2022delta}. Up to now, various \PETs have been proposed, including introducing extra tunable neuron modules~\citep{houlsby2019parameter}, specifying partial parameters to be tunable~\citep{ben2021bitfit} and re-parameterizing part of existing modules in PLMs~\citep{hu2021lora}, etc. \PETs extensively reduce the number of tunable parameters, and still achieves comparable downstream performance to fine-tuning.

Despite the success of \PETs, the mechanism behind it remains unclear. An essential question is: how could the PLM adaptation using different \PETs relate to each other? To answer this question, a direct exploration of the connections among different \PETs is needed, but this would run into a problem: due to the versatile designs of \PETs, the parameter space of various \PETs is inherently different. To address the issue and investigate the above research question, we hypothesize that the adaptations of different \PETs could be re-parameterized as low-dimensional optimizations in a \textbf{unified optimization subspace}. In this sense, optimizing various \PETs can all be viewed as finding optimal solutions within the same subspace. Our hypothesis is inspired by recent findings that despite owning huge amounts of parameters, PLMs have an extremely low intrinsic dimension~\citep{aghajanyan-etal-2021-intrinsic,qin2021exploring}. In this regard, optimizing a certain PET, which is typically a high-dimensional optimization problem, could be equivalently re-parameterized as a low-dimensional optimization problem, while achieving non-trivial performance.

To find evidence for our hypothesis, we design an analysis pipeline as follows: we first independently obtain solutions for different \PETs on a set of tasks. Then we learn to project these solutions to a desired subspace. Meanwhile, we also define a mapping from the subspace to each \PET's original space. We contend that if the found subspace is indeed shared among various \PETs, then two conditions should be satisfied: (1) the optimizations of different \PETs could be equivalently conducted in the found subspace and achieve non-trivial performance, and (2) the local optima of various \PETs have a substantial intersection in the subspace, which means the solution obtained in the subspace using a certain \PET could be directly transferred to other \PETs. If both conditions are well-established for the found subspace, then we could validate the existence of the unified optimization subspace for \PETs.

We conduct experiments on a series of representative NLP tasks, and demonstrate that in the found subspace:

\begin{itemize} %[topsep=1pt, partopsep=1pt, leftmargin=12pt, itemsep=-3pt]
    \item \textbf{Solutions are transferable.} The solution of a \PET in the found subspace not only achieves comparable performance to that in its original \PET space, but can be directly transferred to another \PET, achieving non-trivial performance. 
    \item \textbf{Local optima of \PETs greatly overlap.} When visualizing the performance landscape, we find that there exists a substantial region where different \PETs all perform well, indicating the close connections among different \PETs. 
    \item \textbf{Fine-tuning has strong connection with \PETs.} We extend the above analysis to fine-tuning and show the strong connections between fine-tuning and \PETs.
\end{itemize}

In general, our study is the first work to reveal the connections among different \PETs and fine-tuning from the perspective of subspace optimization, and uncovers the underlying mechanism of PLMs' downstream adaptation. We believe many applications such as the ensemble and transfer among various \PETs can be well empowered by the unified optimization subspace. Our findings can be of interest to researchers who are working on designing better \PETs, and may provide some guidance for using \PETs in many real-world scenarios.
\section{Background}
\paragraph{Delta Tuning.} \PET has been regarded as the new paradigm for PLM adaptation. By training lightweight parameters, \PET yields a compact and extensible model, and could achieve comparable performance to full-parameter fine-tuning. Up to now, various \PET designs have sprung up. For instance, some introduce additional tunable modules after the feed-forward and attention modules in a PLM~\citep{houlsby2019parameter,pfeiffer-etal-2021-adapterfusion}; others prepend tunable prompt tokens into each attention layer \citep{li2021prefix} or only the embedding layer \citep{lester-etal-2021-power}. Another line of work re-parameterizes existing modules with low-rank decompositions~\citep{hu2021lora}. Recently, researchers demonstrate that existing \PET algorithms can be combined simultaneously and achieve better performance~\citep{he2021towards,mao2021unipelt}.

To fathom the mechanisms behind \PET, \citet{he2021towards} pioneered to explore the connections among different \PETs. 
They formalize various \PETs as different ways to compute the modifications on the hidden states and unify different \PETs in terms of \textit{formulas}. However, the unification in the formula does not reveal the essence of \PETs' success, and does not indicate that their internal mechanisms are unified. Our paper differs from theirs in that we explore whether \PETs can be unified in terms of internal mechanisms through the lens of \textit{optimization}. Specifically, we investigate whether the optimization of different \PETs can be unified in a certain subspace.

\paragraph{Intrinsic Dimension.} Intrinsic dimension~\citep{li2018measuring} estimates the minimum number of tunable parameters needed to reach a satisfying performance for neural networks. Instead of training networks in their native parameter space, they linearly re-parameterize all the tunable parameters $\theta_0$ in a randomly oriented subspace: $\theta \leftarrow \theta_0 + \texttt{Proj}(\theta_\text{I})$, where $\texttt{Proj}: \mathbb{R}^{|\theta_\text{I}|} \rightarrow \mathbb{R}^{|\theta_0|}$ denotes a random projection ($|\theta_\text{I}| \ll |\theta_0|$). During optimization, only the low-dimensional vector $\theta_\text{I}$ is tuned. Considering that $|\theta_0|$ could be extremely large, making computation of the projection intractable, \citet{aghajanyan-etal-2021-intrinsic} reduce the computational complexity using Fastfood transformation~\citep{le2013fastfood}. In experiments, they find that for PLMs, a low-dimensional (e.g., $|\theta_\text{I}| \sim 10^3$) re-parameterization could achieve over 85\% performance of fine-tuning ($|\theta_0|$ exceeds millions or even billions). Further, \citet{qin2021exploring} extend the tuning method from fine-tuning to prompt tuning~\citep{lester-etal-2021-power}. They demonstrate that the projection $\texttt{Proj}$ can be trained in order to approximate a better optimization subspace. Based on previous explorations of intrinsic subspace, we aim to validate the existence of a unified subspace for various tuning methods.
\section{Preliminary}
\label{sec:methodology-preliminaries}
Following \citet{he2021towards}, we investigate three representative \PET algorithms to validate our hypothesis, including Adapter~\cite{houlsby2019parameter}, Prefix-tuning~\cite{li2021prefix}, and LoRA~\cite{hu2021lora}. We will first recap the Transformer layer~\citep{vaswani2017attention}, and then give a brief review of the three \PETs. 

\paragraph{Transformer layer.} PLMs generally have multiple Transformer layers, each consisting of a multi-head attention (MHA) and a feed-forward network (FFN). MHA is composed of $\text{N}_h$ attention heads, each containing a query / key / value weight matrix $\mathbf{W}^{(i)}_q$ / $\mathbf{W}^{(i)}_k$ / $\mathbf{W}^{(i)}_v \in \mathbb{R}^{d\times d_h}$, where $d$ denotes the model dimension and $d_h=d/\text{N}_h$. Given a sequence of $n$ vectors $\mathbf{X} \in \mathbb{R}^{n\times d}$, MHA parameterizes them into queries ($\mathbf{Q}^{(i)}$), keys ($\mathbf{K}^{(i)}$) and values ($\mathbf{V}^{(i)}$) as follows:
\begin{align*}
    \mathbf{Q}^{(i)} = \mathbf{X}\mathbf{W}^{(i)}_q, \mathbf{K}^{(i)} = \mathbf{X}\mathbf{W}^{(i)}_k, \mathbf{V}^{(i)} = \mathbf{X}\mathbf{W}^{(i)}_v.
\end{align*}
Each ($\mathbf{Q}^{(i)}$, $\mathbf{K}^{(i)}$, $\mathbf{V}^{(i)}$) triple is then fed into a self-attention function to obtain the $i$-th head's representation $\mathbf{H}_i$. All head representations are then concatenated and combined using an output weight matrix $\mathbf{W}_o \in \mathbb{R}^{d\times d}$:
\begin{align*}
    \mathbf{H}_{i} 
    % &= \text{ATT}(\mathbf{Q}^{(i)}, \mathbf{K}^{(i)}, \mathbf{V}^{(i)}) \\
    &= \text{softmax}(\frac{\mathbf{Q}^{(i)}(\mathbf{K}^{(i)})^T}{\sqrt{d_h}}\mathbf{V}^{(i)}), \\
    \mathbf{H} &= \text{concat}(\mathbf{H}_1,... , \mathbf{H}_{\text{N}_h}) \mathbf{W}_o.
\end{align*}
The FFN module is a two-layer MLP:
\begin{align*}
    \text{FFN}(\textbf{H}) = \sigma(\mathbf{H}\mathbf{W}_1 + \mathbf{b}_1)\mathbf{W}_2 + \mathbf{b}_2,
\end{align*}
where $\mathbf{W}_1 \in \mathbb{R}^{d \times d_m}$, $\mathbf{d} \in \mathbb{R}^{d_m}$, $\mathbf{W}_2 \in \mathbb{R}^{d_m \times d}$ and $\mathbf{b}_2 \in \mathbb{R}^{d}$. $d_m$ is often chosen larger than $d$.

\paragraph{Adapter.} Adapter~\citep{houlsby2019parameter} plugs in light-weight feed-forward networks in Transformer layers (after the MHA module and the FFN module). Every adapter layer typically consists of a down-projection matrix $\mathbf{W}_{\text{down}}\in\mathbb{R}^{d\times r_\text{A}}$, 
a non-linear activation function $f(\cdot)$, and an up-projection matrix 
$\mathbf{W}_{\text{up}}\in\mathbb{R}^{r_\text{A}\times d}$, 
where $r_\text{A}$ denotes the bottleneck dimension. Denote the input as $\mathbf{X}\in\mathbb{R}^{n\times d}$, adapter applies a residual connection as follows:
\begin{align*}
% \label{eq:preliminary-adapter}
    \mathbf{X}\leftarrow \mathbf{X} + f(\mathbf{X}\mathbf{W}_{\text{down}})\mathbf{W}_{\text{up}}.
\end{align*}

\paragraph{Prefix-tuning.}
Prefix-tuning~\citep{li2021prefix} extends the queries $\mathbf{K}^{(i)}$ / the values $\mathbf{V}^{(i)}$ in every MHA module by prepending learnable prefix vectors $\mathbf{P}_\text{K}^{(i)}$ / $\mathbf{P}_\text{V}^{(i)} \in \mathbb{R}^{m \times d_h}$ before them, where $m$ denotes the number of \textit{virtual tokens}. The output of an attention head $\mathbf{H}_i$ can be re-formulated as:
\begin{align*}
% \label{eq:preliminary-prefix}
    \mathbf{H}_i' = \text{ATT}(\mathbf{Q}^{(i)}, [\mathbf{P}_\text{K}^{(i)};\mathbf{K}^{(i)}], [\mathbf{P}_\text{V}^{(i)};\mathbf{V}^{(i)}]),
\end{align*}
where $[\cdot;\cdot]$ denotes concatenation. % We apply prefix tuning to all MHA modules.

\paragraph{LoRA.}
LoRA~\citep{hu2021lora} re-parameterizes the weight updates $\Delta \mathbf{W}$ of the weight matrix $\mathbf{W}$ in the MHA module with low-rank decompositions, i.e., $\Delta \mathbf{W}=\mathbf{W}_{\text{A}}\mathbf{W}_{\text{B}}$, 
where $\mathbf{W}_{\text{A}}\in\mathbb{R}^{d\times r_\text{L}}$ and $\mathbf{W}_{\text{B}}\in\mathbb{R}^{r_\text{L}\times d}$ 
are two learnable low-rank matrices, with $r_\text{L}$ being typically a small integer. For an input $\mathbf{X}\in\mathbb{R}^{n\times d}$, LoRA is formulated as:
\begin{align*}
% \label{eq:preliminary-lora}
    \mathbf{X}\leftarrow \mathbf{X}+s\cdot \mathbf{X}\mathbf{W}_{\text{A}}\mathbf{W}_{\text{B}},
\end{align*}
where $s\ge 1$ is a scaling hyper-parameter. 
% Following~\citet{hu2021lora}, we apply LoRA to the query and value projection matrices in MHA.
\section{Analysis Pipeline}
\label{sec:methodology}

\begin{figure*}[!t]
\centering
\includegraphics[width=0.9\textwidth,trim={1cm 0 0 0}]{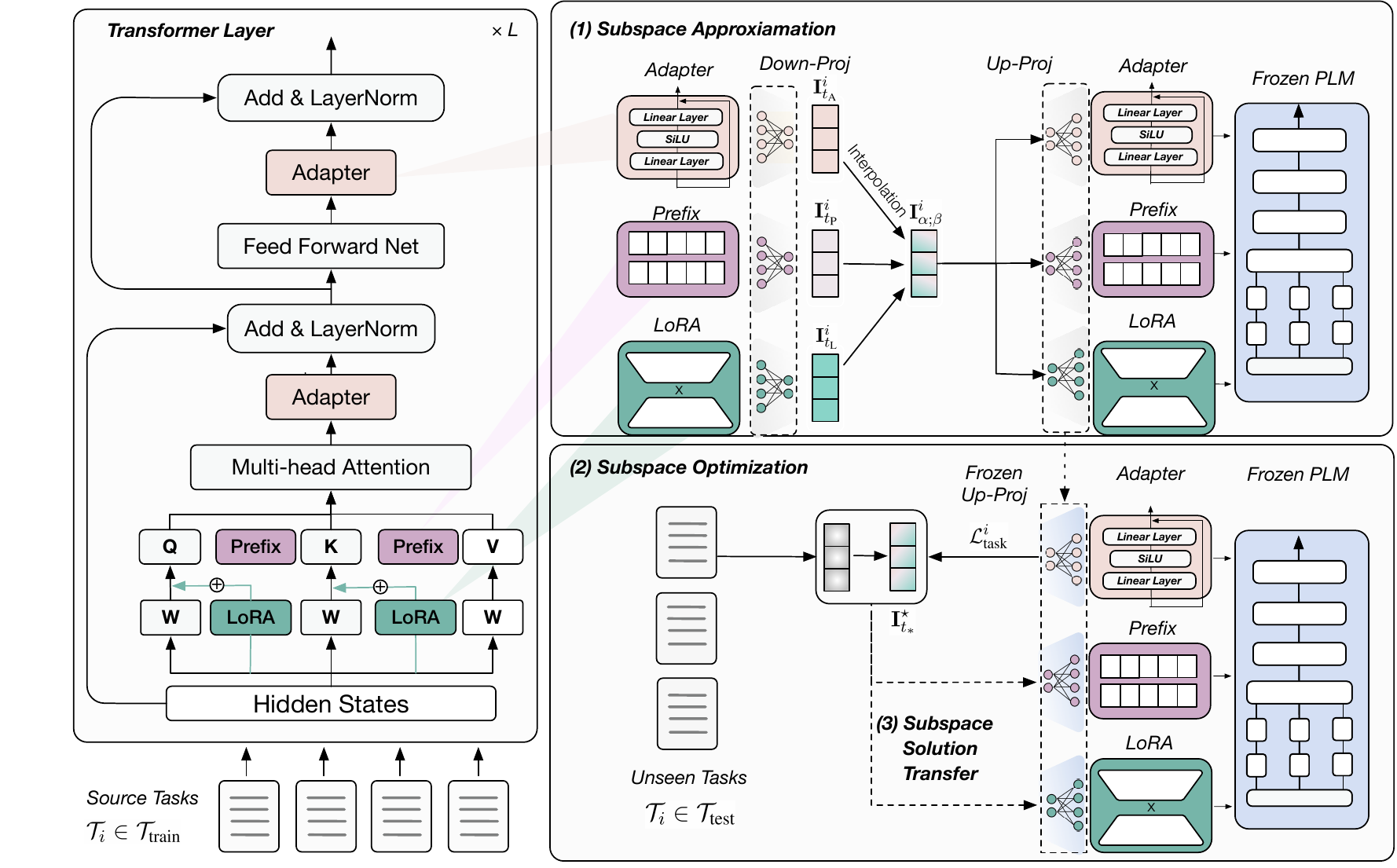}
\caption{Illustration of our analysis pipeline, consisting of (1) \textbf{subspace approximation}, which jointly decomposes \PET solutions into a shared subspace, (2) \textbf{subspace optimization}, which finds subspace solutions for a specific \PET, and (3) \textbf{subspace solution transfer}, which transfers the subspace solution from a source \PET to other \PETs.}
\label{fig:method}
\end{figure*}

As mentioned before, we consider three representative \PETs: Adapter ($t_\text{A}$), Prefix-tuning ($t_\text{P}$), and LoRA ($t_\text{L}$). Each \PET $t_*$ defines a set of tunable parameters $\theta_{t_*}$. To adapt a PLM to a specific downstream task $\mathcal{T}_i$, we optimize $\theta_{t_*}^i$ to minimize the loss function $\mathcal{L}_{\text{task}}^i(\theta_{t_*}^i | \theta_0)$ defined by $\mathcal{T}_i$, where $\theta_0$ denotes the pre-trained weights. To verify our hypothesis that there exists a unified optimization subspace where all \PETs can achieve non-trivial performance, we propose a three-stage analysis pipeline (visualized in Figure~\ref{fig:method}), where the first stage is designed to approximate the desired subspace, so that in the second stage, the optimizations for different \PETs could all be conducted in 
this subspace. This makes it possible to explore the connections of different \PETs in the third stage. Following \citet{qin2021exploring}, to validate the generality of the found subspace and avoid information leakage, we approximate the subspace with a series of training tasks $\mathcal{T}_{\text{train}}$, and conduct subsequent subspace optimization on unseen tasks $\mathcal{T}_{\text{test}}$.

\paragraph{Subspace Approximation.}
To approximate the desired subspace, we decompose and then reconstruct \textit{independent} \PET solutions of $\mathcal{T}_{\text{train}}$. We first train \PETs in their original space, and for each task $\mathcal{T}_i \in \mathcal{T}_{\text{train}}$, we obtain three independent solutions: $\theta_{t_\text{A}}^i$, $\theta_{t_\text{P}}^i$, and $\theta_{t_\text{L}}^i$. Then we assign a down-projection $\texttt{Proj}_{t_*}^{\downarrow}: \mathbb{R}^{|\theta^i_{t_*}|}\rightarrow\mathbb{R}^y$ and an up-projection $\texttt{Proj}_{t_*}^{\uparrow}: \mathbb{R}^y\rightarrow\mathbb{R}^{|\theta^i_{t_*}|}$ for each \PET $t_*$, where $y$ is the dimension of the intrinsic subspace. In practice, both down-projection and up-projection are MLP layers. Each down-projection decomposes a \PET solution into a low-dimensional intrinsic vector $\mathbf{I}_{t_*}^i\in\mathbb{R}^y$:
\begin{align*}
% \label{eq:method-downprojection}
   \mathbf{I}_{t_*}^i = \texttt{Proj}_{t_*}^{\downarrow}(\theta_{t_*}^i).
\end{align*}
Three intrinsic vectors $\mathbf{I}_{t_\text{A}}^i$, $\mathbf{I}_{t_\text{P}}^i$, $\mathbf{I}_{t_\text{L}}^i$ represent different local minima of $\mathcal{T}_i$ in the same subspace. Ideally, if three \PETs can be unified in the subspace, then each vector $\mathbf{I}_{t_*}^i$ could be used to reconstruct any \PET solution ($\theta_{t_\text{A}}^i$, $\theta_{t_\text{P}}^i$, or $\theta_{t_\text{L}}^i$). Therefore, to approximate such a subspace, we facilitate the interaction among different \PETs efficiently by dynamically sampling two random ratios $\alpha\in [0,1], \beta\in [0, 1-\alpha]$, and computing an interpolation of three intrinsic vectors of $\mathcal{T}_i$:
\begin{align*}
   \mathbf{I}_{\alpha; \beta}^i = \alpha \cdot \mathbf{I}_{t_\text{A}}^i + \beta \cdot \mathbf{I}_{t_\text{P}}^i + (1-\alpha-\beta) \cdot \mathbf{I}_{t_\text{L}}^i.
\end{align*}
The interpolation is mapped by each up-projection $\texttt{Proj}_{t_*}^{\uparrow}$ to reconstruct the task solution for each \PET by minimizing the following loss function:
\begin{align*}
% \label{eq:method-distance-loss}
    \mathcal{L}_{\text{dist}}^i (\overline{\theta_{t_*}^i}) = ||\overline{\theta_{t_*}^i} - \theta_{t_*}^i||^2, \quad
    \overline{\theta_{t_*}^i} = \texttt{Proj}_{t_*}^{\uparrow}(\mathbf{I}_{\alpha; \beta}^i).
\end{align*}
To properly guide the reconstructed $\overline{\theta_{t_*}^i}$ to solve task $\mathcal{T}_i$, we also incorporate the original task loss $\mathcal{L}_{\text{task}}^i$. The overall training objective can be formulated as follows:
\begin{align*}
% \label{eq:subspace-approximation-pet-loss}
    \mathcal{L}_{\text{pet}} = \sum_{i=1}^{|\mathcal{T}_{\text{train}}|} \sum_{t_* \in \{t_\text{A}, t_\text{P}, t_\text{L}\}}
    \mathcal{L}_{\text{dist}}^i (\overline{\theta_{t_*}^i}) + \mathcal{L}_{\text{task}}^i(\overline{\theta_{t_*}^i} | \theta_0).
\end{align*}
During this stage, only the down-projections and up-projections are optimized, and other parameters are kept frozen. When this stage finishes, the two projections can be seen as mappings between the unified subspace and each \PET's original space.

\paragraph{Subspace Optimization.}
In the second stage, we investigate whether the optimization in the subspace could achieve comparable performance to the optimization in the original space for unseen tasks $\mathcal{T}_{\text{test}}$. If this holds, then we could empirically validate that the optimizations of different \PETs could be equivalently mapped in this subspace with a low level of error, and it is possible to explore the connections among \PETs in the next stage.

Specifically, we only retain the up-projection $\texttt{Proj}_{t_*}^{\uparrow}$ trained in the first stage. $\texttt{Proj}_{t_*}^{\uparrow}$ defines the mapping from the found subspace to the original \PET space. We keep both PLM and $\texttt{Proj}_{t_*}^{\uparrow}$ frozen during subspace optimization. After that, for each task $\mathcal{T}_i \in \mathcal{T}_{\text{test}}$, the optimization of $t_*$ can be conducted within the subspace defined by $\texttt{Proj}_{t_*}^{\uparrow}$ by merely tuning a randomly initialized intrinsic vector $\mathbf{I}_{t_*}$, which is formulated as:
% In stage 2, we train a certain PET on an unseen dataset $\tilde{\mathcal{D}}$, with its hyper-network being frozen.
\begin{align*}
    \mathbf{I}_{i,t_*}^{\star}=\argmin_{\mathbf{I}_{t_*}^i}\mathcal{L}_{\text{task}}^i(\texttt{Proj}_{t_*}^{\uparrow}(\mathbf{I}_{t_*}^i)|\theta_0).
\end{align*}
% We omit the dependence of objective on input and label for brevity.

\paragraph{Subspace Solution Transfer.}
If the found subspace is shared among different \PETs, then the found solution $\mathbf{I}_{i,t_*}^\star$ in the subspace could be directly transferred to another \PET and achieve non-trivial performance. Taking the transferring between $t_\text{A}$ and $t_\text{P}$ as an example, for a task $\mathcal{T}_i \in \mathcal{T}_{\text{test}}$, we first conduct subspace optimization for $t_\text{A}$ and obtain a well-tuned intrinsic vector $\mathbf{I}_{i,t_\text{A}}^\star$. Then we directly transfer $\mathbf{I}_{i, t_\text{A}}^\star$ to $t_\text{P}$ utilizing its up-projection $\texttt{Proj}_{t_\text{P}}^{\uparrow}$, and obtain a $t_\text{P}$'s solution $\theta_{i, t_\text{A} \rightarrow t_\text{P}}$ in the original \PET space:
\begin{align*}
    \theta_{i, t_\text{A} \rightarrow t_\text{P}} = \texttt{Proj}_{t_\text{P}}^{\uparrow}(\mathbf{I}_{i, t_\text{A}}^\star).
\end{align*}
\section{Experiment}
\label{sec:experiment}

We conduct experiments on representative NLP tasks. We first introduce the experimental setups in~\cref{sec:experiment-setting}, next we approximate the subspace and present the analysis in~\cref{sec:experiment-results}. Lastly, we explore the connection between \PETs and fine-tuning in~\cref{sec:experiment-finetuning}.

\subsection{Experimental Setups}
\label{sec:experiment-setting}

\paragraph{Training Setups.}
We conduct experiments with both single-task and multi-task settings.

In the \textit{single-task setting}, we approximate the unified optimization subspace using only one dataset, i.e., $|\mathcal{T}_{\text{train}}|\!=\!1$. Then we perform the subspace optimization and subspace solution transfer on unseen tasks. However, the subspace approximated with only one task may not generalize well to diverse unseen tasks~\citep{qin2021exploring}. For example, the subspace approximated using a NLI task can hardly be generalized to a QA task. Therefore, for the single-task setting, we only evaluate the found subspace using the unseen tasks belonging to the same category of $\mathcal{T}_{\text{train}}$. 

Besides, we also experiment on the \textit{multi-task setting}, where the unified subspace is approximated with diverse training tasks, i.e., $|\mathcal{T}_{\text{train}}|>1$. The unified \PET subspace found in the multi-task setting is expected to generalize to more diverse tasks than that in the single-task setting.

During subspace solution transfer, we choose the subspace solution that achieves the best transferring performance using the development set, and report its performance on the test set.

\paragraph{Tasks and Datasets.}

In the single-task setting, we experiment with $6$ types of tasks, including:
\begin{itemize} %[topsep=1pt, partopsep=1pt, leftmargin=12pt, itemsep=-3pt]
    \item \textbf{Sentiment Analysis (SA)}: SST-2~\citep{socher-etal-2013-recursive}, Rotten Tomatoes~\citep{pang-lee-2005-seeing}, and Amazon Review~\citep{mcauley2013hidden}.
    \item \textbf{Natural Language Inference (NLI)}: SciTail~\citep{Khot2018SciTaiLAT}, MNLI~\citep{williams-etal-2018-broad}, and RTE~\citep{dagan2005pascal}.
    \item \textbf{Text Classification (TC)}: WiC~\citep{pilehvar-camacho-collados-2019-wic}, and WSC~\citep{levesque2012winograd}.
    \item \textbf{Paraphrase Detection (PD)}: QQP\href{https://quoradata.quora.com/First-Quora-Dataset-Release-Question-Pairs}{(link)}, and MRPC~\citep{dolan-brockett-2005-automatically}.
    \item \textbf{Long-form QA (LF-QA)}: ELI5-ELI5, ELI5-Askh, and ELI5-Asks~\citep{fan-etal-2019-eli5}.
    \item \textbf{Multiple-choice QA (MC-QA)}: CoPA~\citep{gordon-etal-2012-semeval}, DREAM~\citep{saha-etal-2018-duorc}, QuaRTz~\citep{tafjord-etal-2019-quartz} and CODAH~\citep{chen-etal-2019-codah}.
\end{itemize}

We include more diverse datasets in the multi-task setting, and randomly partition them into $60$ training tasks $\mathcal{T}_\text{train}$ and $9$ test tasks $\mathcal{T}_\text{test}$. More details are left in \cref{sec:appendix-datasets}.

\paragraph{Evaluation Metrics.}
For each dataset, we use the common evaluation metric, e.g., \textsc{Rouge-L} for LF-QA , \textsc{F1} for SA and NLI, \textsc{accuracy} for TC, PD and MC-QA. Denote $E_{\text{ori}}$ as the performance achieved by \PET in the original space, and $E_{\text{sub}}$ as the performance achieved by optimization within the subspace, we report the relative recovering performance (\%), i.e., $\frac{E_{\text{sub}}}{E_{\text{ori}}}$ in all experiments.

\paragraph{Models.}
We use $\text{T5}_{\texttt{BASE}}$~\cite{2020t5} as the backbone model, and unify all tasks into a text-to-text format without loss of generality. We set the dimension of the subspace to $4$ in single-task setting and $100$ in multi-task setting. During subspace optimization, only $4$ or $100$ free parameters are tuned, compared with $220$M for fine-tuning. We choose the intrinsic dimension according to our preliminary experiment. The single-task performances of different intrinsic dimensions in $\{4, 8, 16\}$ do not vary much. The multi-task performance gets better when the intrinsic dimension increases. Practically, we find a dimension of 100 strikes a satisfying balance between performance and computational resources. The details of implementation are shown in~\cref{sec:appendix-hyperparameters}.
% \footnote{Text-to-text is a unified formation for almost all textual tasks, and has been widely used in previous works~\citep{ding2022delta,ye2021crossfit}. Formatting a task into text-to-text format does not degrade the performance of the generative models such as the T5 we used~\citep{ding2022delta}}

% \input{tables/multi_task_result}

\subsection{Experimental Results}
\label{sec:experiment-results}

\subsubsection{Single-task Setting}
\label{sec:experiment-results-singletask}
\begin{table}[t!]
    \centering
    \scriptsize
    \setlength{\tabcolsep}{5pt}
    \begin{tabular}{l l *{4}{r}}
    \toprule
        $\mathcal{T}_{\text{train}}$ & $\mathcal{T}_{\text{test}}$ & \textbf{Adapter} & \textbf{LoRA} & \textbf{Prefix} & \textbf{Avg.} \\
        \midrule
        \multirow{2}{*}{SST-2} & Rotten Tomatoes & 101.8 & 100.1 & 99.3 & 100.4 \\
         & Amazon Review & 98.0 & 96.9 & 98.2 & 97.7 \\
        \midrule
        \multirow{2}{*}{MNLI} & SciTail & 82.9 & 79.8 & 84.4 & 82.4 \\
        & RTE & 95.4 & 68.2 & 80.2 & 81.3 \\
        \midrule
        \multirow{1}{*}{WiC} & WSC & 70.6 & 57.6 & 77.1 & 68.4 \\
        \midrule
        \multirow{1}{*}{QQP} & MRPC & 85.4 & 84.3 & 83.1 & 84.3 \\
        \midrule
        \multirow{2}{*}{ELI5-ELI5} & ELI5-Askh & 91.4 & 87.6 & 80.1 & 86.4 \\
        & ELI5-Asks & 96.6 & 95.0 & 94.9 & 95.5 \\
        \midrule
        \multirow{3}{*}{DREAM} & CODAH & 77.4 & 70.4 & 74.0 & 73.9 \\
        & QuaRTz & 75.6 & 78.5 & 74.7 & 76.3 \\
        & CoPA & 98.3 & 71.6 & 92.5 & 87.5 \\
        \midrule
        & \textbf{Avg.} & 88.5 & 80.9 & 85.3 & 84.9 \\
        \bottomrule
    \end{tabular}
    \caption{Relative performance (\%) for subspace optimization under the single-task setting.}
    \label{tab:single-task-subspace-optimization}
\end{table}

\paragraph{Subspace Optimization.} The results of subspace optimization are presented in~\cref{tab:single-task-subspace-optimization}. On average, for all three \PETs, optimization within the subspace can recover more than $80$\% performance of the original space. Among three \PETs, Adapter achieves the best recovering performance ($\approx 90$\%), despite only tuning 4 free parameters. This indicates that we have found a satisfying optimization subspace that could recover most of the performance of the original space\footnote{\citet{aghajanyan-etal-2021-intrinsic} deem $85$\% as a satisfying recovering performance for an intrinsic subspace. Although the performance of our method could be a bit lower under certain cases, we contend that the performance is already non-trivial.}, and the subspace can be generalized to unseen tasks $\mathcal{T}_{\text{test}}$ belonging to the same category of $\mathcal{T}_{\text{train}}$.

\paragraph{Subspace Solution Transfer.} Then we transfer the solution found with a source \PET to other \PETs. The results are presented in~\cref{tab:single-task-solution-transfer}. On $6$ out of the $11$ tasks, transferring the subspace solution from a source \PET to a target \PET achieves more than 80\% recovering performance, and achieves $82.6$\% on average across all tasks. This demonstrates that the transferred intrinsic vector yields \PETs with non-trivial performance. In particular, in the category of sentiment analysis, the subspace of three \PETs approximated on SST-2 serves as an excellent optimization subspace for similar tasks (R. Tomatoes and A. Review). Performing optimization in this subspace with an arbitrary source \PET and transfer the found intrinsic vector to other \PETs yield performance comparable to or even surpass the original \PET space.

\begin{table}[t!]
    \centering
    \scriptsize
    \setlength{\tabcolsep}{1pt}
    \begin{tabular}{l l *{7}{r}}
    \toprule
        $\mathcal{T}_{\text{train}}$ & $\mathcal{T}_{\text{test}}$ & $\mathbf{A} \!\!\rightarrow\!\! \mathbf{L}$ & $\mathbf{A} \!\!\rightarrow\!\! \mathbf{P}$ & $\mathbf{L} \!\!\rightarrow\!\! \mathbf{A}$ & $\mathbf{L} \!\!\rightarrow\!\! \mathbf{P}$ & 
        $\mathbf{P} \!\!\rightarrow\!\! \mathbf{A}$ & 
        $\mathbf{P} \!\!\rightarrow\!\! \mathbf{L}$ & \textbf{Avg.} \\
        % \cmidrule(lr){3-4}\cmidrule(lr){5-6}\cmidrule(lr){7-8}
        % & & \multicolumn{1}{c}{LoRA} & \multicolumn{1}{c}{Prefix} & \multicolumn{1}{c}{Adapter} & \multicolumn{1}{c}{Prefix} & \multicolumn{1}{c}{Adapter} & \multicolumn{1}{c}{LoRA} & \\
        \midrule
        \multirow{2}{*}{SST-2} & R. Tomatoes & 100.6 & 99.0 & 100.7 & 98.8 & 101.0 & 100.8 & 100.2 \\
        & A. Review & 97.1 & 97.7 & 97.7 & 98.1 & 97.6 & 96.7 & 97.5 \\
        \midrule
        \multirow{2}{*}{MNLI} & SciTail & 81.7 & 83.0 & 83.2 & 83.6 & 83.4 & 80.9 & 82.6 \\
        & RTE & 62.7 & 74.3 & 81.7 & 78.2 & 55.0 & 80.0 & 72.0 \\
        \midrule
        \multirow{1}{*}{WiC} & WSC & 72.7 & 54.3 & 58.8 & 57.1 & 76.5 & 69.7 & 64.9 \\
        \midrule
        \multirow{1}{*}{QQP} & MRPC & 83.7 & 65.7 & 83.7 & 69.1 & 84.8 & 85.4 & 78.7 \\
        \midrule
        \multirow{2}{*}{ELI5-ELI5} & ELI5-Askh & 88.0 & 79.4 & 91.3 & 78.1 & 90.3 & 87.0 & 85.7 \\
        & ELI5-Asks & 95.9 & 95.4 & 91.3 & 92.5 & 97.8 & 96.1 & 94.8 \\
        \midrule
        \multirow{3}{*}{DREAM} & QuaRTz & 76.2 & 71.9 & 76.0 & 74.5 & 76.4 & 77.2 & 75.4 \\
        & CODAH & 63.0 & 61.6 & 74.3 & 58.1 & 83.9 & 69.0 & 68.3 \\
        & CoPA & 77.3 & 100.3 & 96.0 & 103.1 & 91.7 & 65.6 & 89.0 \\
        \midrule
        & \textbf{Avg.} & 81.7 & 80.2 & 85.0 & 81.0 & 85.3 & 82.6 & 82.6\\
        \bottomrule
    \end{tabular}
    \caption{Relative performance (\%) for subspace solution transfer under the single-task setting. $\mathbf{A}$, $\mathbf{P}$, and $\mathbf{L}$ refer to Adapter, Prefix-tuning, and LoRA, respectively. As an example, $\mathbf{A}\!\!\rightarrow\!\!\mathbf{L}$ means we obtain $\mathbf{I}_{t_\text{A}}^\star$ by conducting subspace optimization with Adapter (source \PET), and then transfer the subspace solution to LoRA (target \PET) with the fixed up-projection, i.e., $\texttt{Proj}_{t_\text{P}}^{\uparrow}(\mathbf{I}_{t_\text{A}}^\star)$.}
    % R.Tomatoes refers to Rotten Tomatoes, and A. Reivew refers to Amazon Review.}
    \label{tab:single-task-solution-transfer}
\end{table}

However, we also observe that the transferred \PET on WSC does not perform very well, achieving only $64.9$\% recovering performance. We argue that this may be due to the inherent difference between WiC ($\mathcal{T}_\text{train}$) and WSC ($\mathcal{T}_\text{test}$): WiC evaluates the quality of context-sensitive representations, while WSC is a coreference resolution task, which requires slightly distinct language skills from WiC. Besides, the performances of the transferred \PET on DREAM are also slightly below the expectation, this may due to the domain differences between DREAM and the target tasks. Although they all belong to multi-choice QA, their domains differ significantly. In fact, these unwanted transferred performance can be substantially improved in multi-task setting, as we will see in the next section.

In addition, we do not find a significant difference in the transferability among different \PETs. In general, when serving as the source \PET, Adapter has slightly worse transferability than other two \PETs. Besides, the transferability of a \PET seems to have a weak correlation to its performance of subspace optimization.

\subsubsection{Multi-task Setting}
\label{sec:experiment-results-multitask}
To improve the subspace's task-level generalization, we propose to approximate the subspace in a multi-task manner ($60$ training tasks in total), and test the generalization ability of the approximated subspace on $6$ categories of unseen tasks. Note for the multi-task setting, the subspace optimization and subspace solution transfer are carried out within the same subspace for \textit{all} the unseen tasks.

\begin{table}[t!]
    \centering
    \scriptsize
    \begin{tabular}{l *{4}{r}}
    \toprule
        $\mathcal{T}_{\text{test}}$ & \textbf{Adapter} & \textbf{LoRA} & \textbf{Prefix} & \textbf{Avg.} \\
        \midrule
        Rotten Tomatoes & 99.7 & 98.2 & 100.3 & 99.4\\
        Yelp Polarity & 99.5 & 99.5 & 98.7 & 99.2\\
        \midrule
        WSC & 88.2 & 75.8 & 80.0 & 81.3\\
        \midrule
        AI2 ARC & 93.2 & 87.9 & 78.8 & 86.6\\
        QASC & 99.3 & 71.4 & 90.9 & 87.2\\
        QuaRTz & 96.6 & 86.9 & 77.3 & 86.9\\
        \midrule
        BLiMP-ANA & 100.0 & 100.0 & 51.0 & 83.7\\
        \midrule
        ELI5-Asks & 99.9 & 99.6 & 94.3 & 97.9\\
        \midrule
        ETHOS-Gender & 79.6 & 88.9 & 59.0 & 75.8\\
        \midrule
        \textbf{Avg.} & 95.1 & 89.8 & 81.1 & 88.7\\
        \bottomrule
    \end{tabular}
    \caption{Relative performance (\%) for subspace optimization under the multi-task setting.}
    \label{tab:multi-task-subspace-optimization}
\end{table}

\paragraph{Subspace Optimization.} 
The results of subspace optimization under the multi-task setting are shown in~\cref{tab:multi-task-subspace-optimization}. In general, three \PETs achieve non-trivial ($88.7$\%) performance during subspace optimization on unseen tasks. Among three \PETs, Adapter still performs the best, achieving $95.1$\% of its original performance. However, the performance of Prefix-tuning is about $10$\% poorer than Adapter and LoRA. We observe that when approximating the subspace, the loss of Prefix-tuning converges much slower than Adapter and LoRA, which may partially explain the poorer performance of Prefix-tuning. We leave further exploration of this phenomenon as future work.

\paragraph{Subspace Solution Transfer.}
The results are presented in~\cref{tab:multi-task-solution-transfer}. On $8$ out of $9$ tasks, \PETs recover around or more than $80$\% their performance in the original space. The non-trivial results demonstrate that (1) for most of the investigated unseen tasks, the local optima found by a source \PET can be directly transferred to a target \PET and achieve non-trivial performance; (2) the subspace approximated with multiple training tasks can be well generalized to diverse unseen tasks. Both findings provide strong evidence for our hypothesis that different \PETs can be re-parameterized into a unified optimization subspace.

\begin{table}[t!]
    \centering
    \scriptsize
    \setlength{\tabcolsep}{2pt}
    \begin{tabular}{l *{10}{r}}
    \toprule
          $\mathcal{T}_{\text{test}}$ &
          $\mathbf{A} \!\!\rightarrow\!\! \mathbf{L}$ & $\mathbf{A} \!\!\rightarrow\!\! \mathbf{P}$ & $\mathbf{L} \!\!\rightarrow\!\! \mathbf{A}$ & $\mathbf{L} \!\!\rightarrow\!\! \mathbf{P}$ & 
        $\mathbf{P} \!\!\rightarrow\!\! \mathbf{A}$ & 
        $\mathbf{P} \!\!\rightarrow\!\! \mathbf{L}$ & \textbf{Avg.} \\
        % \cmidrule(lr){2-3}\cmidrule(lr){4-5}\cmidrule(lr){6-7}
        % & \multicolumn{1}{c}{LoRA} & \multicolumn{1}{c}{Prefix} & \multicolumn{1}{c}{Adapter} & \multicolumn{1}{c}{Prefix} & \multicolumn{1}{c}{Adapter} & \multicolumn{1}{c}{LoRA} & \\
        \midrule
        Rotten Tomatoes& 97.5 & 96.6 & 98.5 & 95.9 & 98.3 & 96.2 & 97.2\\
        Yelp Polarity & 99.1 & 97.5 & 99.4 & 98.1 & 98.5 & 98.4 & 98.5\\
        \midrule
        WSC & 90.9 & 85.7 & 97.1 & 94.3 & 91.2 & 93.9 & 92.2\\
        \midrule
        AI2 ARC & 78.7 & 79.3 & 87.7 & 76.9 & 85.2 & 87.9 & 82.6\\
        QASC & 65.1 & 63.6 & 98.7 & 72.1 & 105.2 & 68.0 & 78.8 \\
        QuaRTz & 77.4 & 71.7 & 90.9 & 73.9 & 83.4 & 78.1 & 79.2 \\
        \midrule
        BLiMP-ANA & 98.0 & 47.0 & 95.0 & 49.0 & 92.0 & 95.0 & 79.3\\
        \midrule
        ELI5-Asks & 90.6 & 89.7 & 89.6 & 95.0 & 95.8 & 87.4 & 91.4\\
        \midrule
        ETHOS-Gender & 55.5 & 57.8 & 53.0 & 73.3 & 68.7 & 69.8 & 63.0\\
        \midrule
        \textbf{Avg.} & 83.6 & 76.5 & 90.0 & 80.9 & 90.9 & 86.1 & 84.7 \\
        \bottomrule
    \end{tabular}
    \caption{Relative performance (\%) for subspace solution transfer under the multi-task setting.}
    \label{tab:multi-task-solution-transfer}
\end{table}

We also observe that, the transferring performance on WSC is far better than that in the single-task setting, demonstrating the benefits of including diverse training tasks in subspace approximation. However, we still find that there are cases where the subspace solution of a source \PET has poor transferability to another \PET. For instance, the transferring performance of different \PETs on ETHOS-Gender is only $63.0$\%. We conjecture that it is due to the gap between ETHOS-Gender and the training tasks $\mathcal{T}_\text{train}$. As demonstrated by \citet{qin2021exploring}, increasing the diversity and number of training tasks could significantly improve the generalization ability of the subspace on unseen tasks. We expect future works to apply our analysis to more diverse training tasks.

\begin{figure*}[t]
     \begin{subfigure}[b]{0.31\textwidth}
         \centering
         \includegraphics[width=\textwidth]{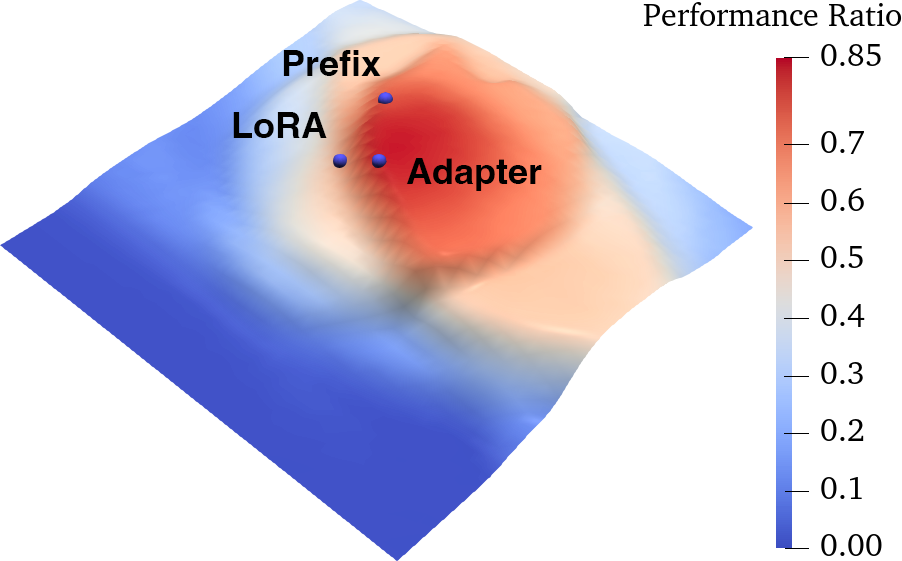}
         \caption{SciTail (single-task)}
         \label{fig:landscape-sctail}
     \end{subfigure}
    %  \hfill
     \begin{subfigure}[b]{0.31\textwidth}
         \centering
         \includegraphics[width=\textwidth]{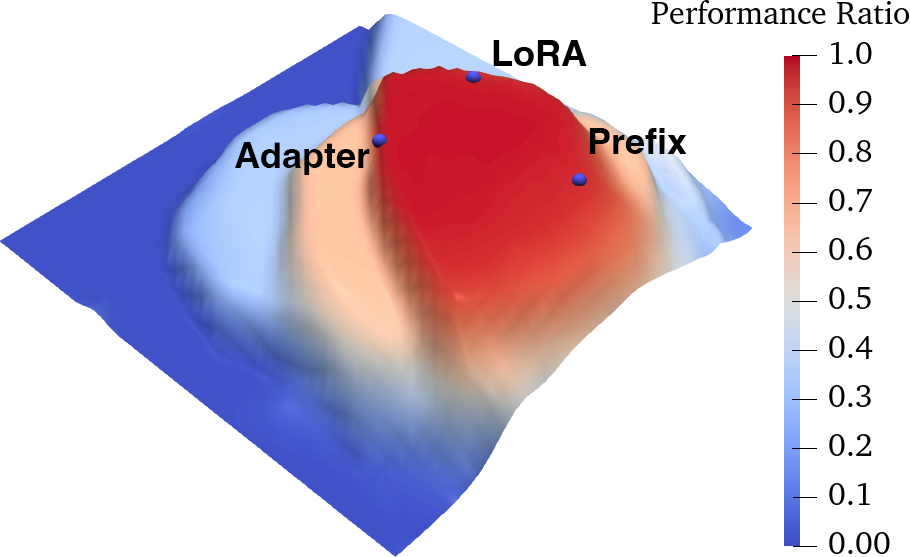}
         \caption{Yelp Polarity (multi-task)}
         \label{fig:landscape-yelp}
     \end{subfigure}
    %  \hfill
     \begin{subfigure}[b]{0.31\textwidth}
         \centering
         \includegraphics[width=\textwidth]{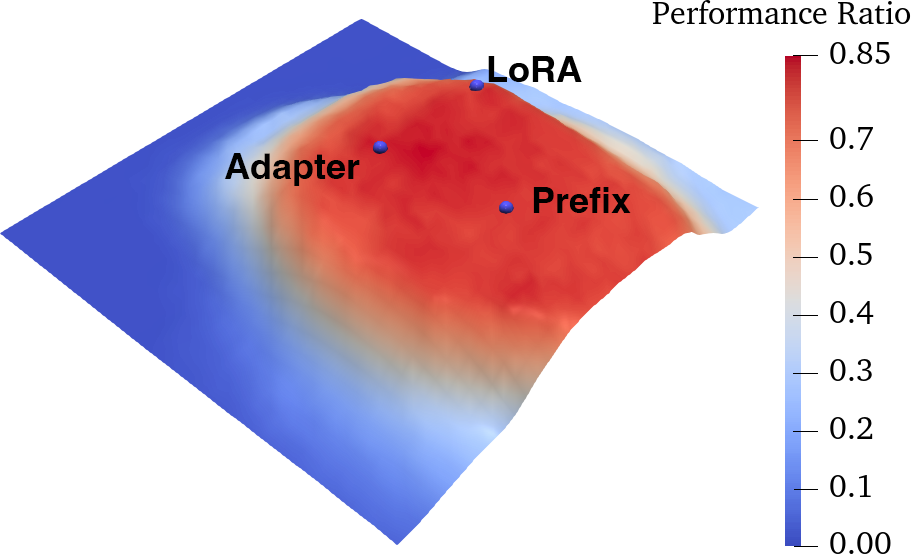}
         \caption{AI2 ARC (multi-task)}
         \label{fig:landscape-ai2}
     \end{subfigure}
        \caption{Performance landscape visualization on three datasets (SciTail, Yelp Polarity, and AI2 ARC). We highlight the subspace solutions ($\mathbf{I}_{t_\text{A}}, \mathbf{I}_{t_\text{L}}$ and  $\mathbf{I}_{t_\text{P}}$) found independently by the three \PETs.}
        \label{fig:landscapes}
\end{figure*}

Furthermore, comparing the transferability of different \PETs, we find that similar to the single-task setting, Adapter is still slightly worse than the other two \PETs. We also find there is no symmetry in the transferability. For example, the average transferring performance from Prefix-tuning to Adapter achieves $90.9$\% of its performance, while the performance in the opposite direction only reaches $76.5$\%.

\subsubsection{Performance Landscape Visualization}
\label{sec:experiment-results-landscape}

From~\cref{tab:single-task-solution-transfer,tab:multi-task-solution-transfer}, we observe non-trivial transferring performance among different \PETs. These results demonstrate that the local optima of different \PETs have a substantial overlap in the approximated subspace for the investigated unseen tasks. It is natural to be concerned about how large this overlap area is since a larger overlap may indicate closer connection of different \PETs. Therefore, for both single-task and multi-task settings, we visualize the performance landscape to understand to what extent the local optima of different \PETs in intrinsic subspace overlap with each other.

Specifically, denote $\mathbf{I}_0$ as an origin, and $\mathbf{u}, \mathbf{v}$ as two orthogonal directions. Let $\alpha$, $\beta$ be two coordinates in the $2$-dimensional space spanned by $\mathbf{u}$ and $\mathbf{v}$. Each solution $\mathbf{I}_0+\alpha\mathbf{u}+\beta\mathbf{v}$ in the subspace can be mapped by the up-projection of \PET $t_*$ to the solution $\texttt{Proj}_{t_*}^{\uparrow} (\mathbf{I}_0+\alpha\mathbf{u}+\beta\mathbf{v})$ in the \PET space. Denote $E(\texttt{Proj}_{t_*}^{\uparrow} (\mathbf{I}_0+\alpha\mathbf{u}+\beta\mathbf{v}))$ as the performance of the recovered \PET, and $E_\text{PET}$ as the average performance of the three \PETs in their original space, we plot the relative performance along these two directions as follows:
\begin{align*}
    \mathcal{P} \!=\! \frac{1}{3}\!\sum_{t_*\in\{t_\text{A}, t_\text{P}, t_\text{L}\}}\! \frac{E(\texttt{Proj}_{t_*}^{\uparrow} (\mathbf{I}_0+\alpha\mathbf{u}+\beta\mathbf{v}))}{E_\text{PET}}.
\end{align*}

If $\mathcal{P}$ is high at ($\alpha$, $\beta$), then it means three \PETs all correspond to high performance at this point. Let $\textbf{I}_{t_\text{A}}, \textbf{I}_{t_\text{P}}, \textbf{I}_{t_\text{L}}$ denote the optimal solution obtained by tuning each \PET in the subspace independently. We visualize the performance landscape around these optimal solutions. Without loss of generality, we choose $\textbf{I}_{t_\text{A}}$ as the origin, and select two orthogonal axes $\mathbf{u}, \mathbf{v}$ as follows:
\begin{align*}
    \textbf{u} \!=\! \frac{\textbf{I}_{t_\text{P}} - \textbf{I}_{t_\text{A}}}{\lVert \textbf{I}_{t_\text{P}} - \textbf{I}_{t_\text{A}}\rVert},
    \tilde{\textbf{v}} \!=\! \textbf{I}_{t_\text{L}} - \textbf{I}_{t_\text{A}},
    \textbf{v} \!=\! \frac{\tilde{\textbf{v}} - \langle \tilde{\textbf{v}}, \mathbf{u}\rangle \mathbf{u}}{\lVert \tilde{\textbf{v}} - \langle \tilde{\textbf{v}}, \mathbf{u}\rangle \mathbf{u}\rVert}.
\end{align*}
We traverse $\alpha$ and $\beta$ from $-4$ to $4$ with a step size of $0.4$. Due to the length limit, we only show in~\cref{fig:landscapes} the performance landscape on (1) SciTail of the single-task setting, (2) Yelp Polarity and AI2 ARC under the multi-task setting.

We observe that the subspace solutions of different \PETs almost lie in the same optimal region for each task. Comparing the landscape of both single-task and multi-task settings, the highland area is much wider in the multi-task setting. This may explain the better transferability under the multi-task setting. In general, the above results demonstrate that the optimal solutions of different \PETs indeed have a large overlap, otherwise there should not exist such a flat performance highland.

\subsection{Extension to Fine-tuning}
\label{sec:experiment-finetuning}

\begin{table}[t!]
    \centering
    \scriptsize
    \begin{tabular}{l *{5}{r}}
    \toprule
         \diagbox{\textbf{Src.}}{\textbf{Tgt.}} & Adapter & Prefix & LoRA & Fine-tune \\
         \midrule
         Adapter & $100.6$ & $100.1$ & $97.7$ & $95.5$ \\
         Prefix & $101.2$ & $100.4$ & $97.7$ & $95.0$ \\
         LoRA & $101.1$ & $100.2$ & $97.5$ & $95.8$ \\
         Fine-tune & $101.1$ & $99.7$ & $97.1$ & $96.1$ \\
         \bottomrule
    \end{tabular}
    \caption{Relative performance (\%) for subspace optimization and subspace solution transfer for different tuning methods on Rotten Tomatoes. The subspace is approximated on SST-2. We transfer the subspace solution from a source tuning method to a target one.}
    \label{tab:finetuning-results}
\end{table}

Finally, we extend our analysis pipeline to fine-tuning, and investigate its connection with \PETs. However, directly training a down-projection and up-projection for fine-tuning encounters difficulty: if we still use an MLP layer (as introduced in~\cref{sec:methodology}) to implement the projections, the number of trainable parameters will be $\mathcal{O}(yN)$, where $N$ is the number of parameters of the PLM, and $y$ is the dimension of our intrinsic subspace. Since PLMs generally contain tremendous parameters, it is intractable to train such an MLP. To alleviate the problem, we turn to using Fastfood transformation~\cite{yang2015deep,aghajanyan-etal-2021-intrinsic} as an alternative. It is an approximation for the linear projection, but requires far fewer parameters. Specifically, the up-projection using Fastfood transformation can be formalized as follows:
\begin{align*}
    \tilde{\theta}=\theta_0+\mathbf{I}M, \quad M=HG\Pi HB,
\end{align*}
where $\tilde{\theta}$ denotes the tunable parameters in the original space, $\theta_0$ is the pre-trained weights, and $\mathbf{I}$ is the intrinsic vector. The Fastfood matrix $M$ can be factorized as a Hadamard matrix $H$, a random permutation matrix $\Pi$, a diagonal matrix $G$ with each element sampled from a standard normal distribution, and a diagonal matrix $B$ with each element being $\pm 1$ with equal probability. Unlike previous work that uses a random and frozen Fastfood matrix~\cite{yang2015deep,aghajanyan-etal-2021-intrinsic}, we optimize the matrix $G$ to better approximate the desired subspace. For \PETs, the projection is implemented the same as before. More detailed implementations are described in \cref{sec:appendix-results-share-intrinsic}.

We perform subspace approximation on SST-2, and report the results of subspace optimization and subspace solution transfer on Rotten Tomatoes. As shown in~\cref{tab:finetuning-results}, we find that: (1) all the \PETs and fine-tuning can achieve satisfactory results in subspace optimization, and the solution found by any source tuning method can be transferred to other target tuning methods and achieve non-trivial performance. This demonstrates the close connection between \PETs and fine-tuning in the approximated subspace. (2) Since the Fastfood transformation is an approximation for linear projection, its representation ability may be limited. Therefore, the transferring performance of fine-tuning is slightly inferior to other \PETs.

In general, the above finding implicates that all tuning methods may be re-parameterized to a unified optimization subspace, which also sheds light on the reason why different \PETs optimize distinct sets of parameters, but all achieve comparable downstream performance to fine-tuning.
\section{Conclusion}
In this work, we explore the hypothesis that the adaptations of different delta tuning methods could all be re-parameterized as low-dimensional optimizations in a unified optimization subspace. The empirical results provide strong evidence for our hypothesis. We also extend our analysis to find the connection between fine-tuning and delta tuning. We hope our findings could provide insights for future research in designing better tuning methods and understanding the mechanisms behind PLM adaptation.

\section*{Acknowledgments}
This work is supported by the National Key R\&D Program of China (No. 2020AAA0106502) and Institute Guo Qiang at Tsinghua University.

Yujia Qin, Weize Chen, and Jing Yi designed the methods and wrote the paper. Jing Yi, WeizeChen, and Yujia Qin conducted the experiments. Yankai Lin, Zhiyuan Liu, Maosong Sun, and Jie Zhou advised the project. All authors participated in the discussion.

\section*{Limitations}

The limitations of this paper are listed as follows:
\begin{itemize} [topsep=1pt, partopsep=1pt, leftmargin=12pt, itemsep=-3pt]
    \item Under some settings, the recovering performance of the subspace optimization and subspace solution transfer can still be improved. % We argue that this is because: (1) the representation ability of the subspace is poorer than the original space, since we tune far fewer parameters; (2) the relation of training tasks $\mathcal{T}_\text{train}$ and test tasks $\mathcal{T}_\text{test}$ has a great impact on the recovering performance. As shown by \citet{qin2021exploring}, increasing the number and diversity of $\mathcal{T}_\text{train}$ could improve the generalization ability of the subspace on unseen tasks. On average, we achieve $82.6$\% and $84.7$\% recovering performance in subspace solution transfer for the single-task and multi-task setting, respectively. Such performance is already non-trivial, and can be seen as strong evidence for our hypothesis.
    \item We only conduct the experiments using $\text{T5}_\texttt{BASE}$ model. We expect future works to apply our analysis pipeline to other kinds of PLMs, and PLMs with larger sizes. % Besides, we consider around $70$ representative NLP tasks in this paper, there are other tasks that we do not evaluate. We leave the exploration on other kinds of tasks as future work.
\end{itemize}

% Due to the different structure of PETs, we cannot directly study their connection in the original optimization space. Therefore, we resort to the intrinsic subspace, and try to unify different \PETs within the same optimization subspace. Moreover, we flatten the parameters of a PET and study it as a whole. Different position of PETs in the transformer still need to be studied.

% \section*{Ethics Statement}

\bibliography{anthology,custom,ipt}
\bibliographystyle{acl_natbib}

\newpage
\appendix
\section{Appendices}
\label{sec:appendix}
\subsection{Datasets}
\label{sec:appendix-datasets}
All the datasets used in the multi-task setting are listed in Table~\ref{tab:multitask_train} and Table~\ref{tab:multitask_test}. All these datasets are downloaded from Huggingface Datasets~\citep{lhoest2021datasets}.

\subsection{Hyper-parameters and Network Structure for~\cref{sec:experiment-results}}
\label{sec:appendix-hyperparameters}

The down-projection $\texttt{Proj}_{t_*}^{\downarrow}$ is a two-layer MLP, with the first linear layer $f_\text{down-1}:\mathbb{R}^{|\theta_{t_*}|}\rightarrow \mathbb{R}^y$, and the second linear layer $f_\text{down-2}:\mathbb{R}^y\rightarrow \mathbb{R}^y$, where $y$ is the dimension of intrinsic subspace. We use \texttt{tanh} as the activation function between $f_\text{down-1}$ and $f_\text{down-2}$. The up-projection $\texttt{Proj}_{t_*}^{\uparrow}$ is a single linear layer $f_\text{up}:\mathbb{R}^y\rightarrow \mathbb{R}^{|\theta_{t_*}|}$. Note for all the linear layers of the projections, we do not include the bias term.

To ensure that the number of parameters is consistent across the three \PETs, we set $r_\text{A}$ as $12$, $r_\text{L}$ as $10$, ${m}$ as $120$, and $d_\text{P}$ as $24$. $d_\text{P}$ refers to the hidden dimension of the two-layer MLP that is used to re-parameterize Prefix vectors $\mathbf{P}_\text{K}^{(i)}$ / $\mathbf{P}_\text{V}^{(i)}$, see~\citealt{li-liang-2021-prefix} for more details. The meanings of other notations are the same as those in~\cref{sec:methodology-preliminaries}. In this way, the number of parameters of $\theta_{t_\text{A}}^i$, $\theta_{t_\text{P}}^i$ and $\theta_{t_\text{L}}^i$ are all $1105920$. Moreover, following ~\citealt{Houlsby2019Adapter}, we choose a SiLU activation function for Adapter. Following ~\citealt{Hu2021LoRALA}, we set the scaling factor $s$ in LoRA as $1.6$. In our implementation, LoRA is applied to $\mathbf{Q}$ and $\mathbf{V}$ matrices in the MHA module.

During subspace approximation, for the multi-task setting, we randomly sample $20000$ instances from the original training set of each dataset, and blend them together to form the multi-task training set. Similarly, we sample another $240$ instances from each dataset to form the validation set. We set the learning rate as $1\times10^{-4}$, batch size as $4$. We train the model for $1$ epoch and evaluate on validation set for every $1000$ steps. For the single-task setting, we perform grid search using the learning rates in \{$1\times10^{-5}$, $5\times10^{-5}$\} and set the batch size as $8$. We train the model for a maximum step of $100000$, and evaluate on validation set for every $1000$ steps. When conducting subspace optimization, we perform grid search using the learning rate in \{$1\times10^{-2}$, $5\times10^{-2}$\} and set the batch size as $8$. We train the model for a maximum of $5000$ steps and evaluate on validation set every $500$ steps.

We train the model using Adafactor~\citep{shazeer2018adafactor} with a constant learning rate in all experiments. Intrinsic dimension $y$ (the dimension of the approximated subspace) is set to $4$ in single-task setting and $100$ in multi-task setting. We additionally set a ratio $\alpha = 10$ to balance the reconstruction loss $\mathcal{L}_{\text{dist}}^i (\overline{\theta_{t_*}^i})$ and original task loss $\mathcal{L}_{\text{task}}^i(\overline{\theta_{t_*}^i} | \theta_0)$, i.e., $\mathcal{L}_{\text{pet}} = 10 * \mathcal{L}_{\text{dist}}^i (\overline{\theta_{t_*}^i}) + \mathcal{L}_{\text{task}}^i(\overline{\theta_{t_*}^i} | \theta_0)$.
All experiments are carried out on NVIDIA 32GB V100 GPU.

\subsection{Simplification of Subspace Approximation}
\label{sec:appendix-shared-intrinsic}

\begin{table}[t!]
    \centering
    \scriptsize
    \setlength{\tabcolsep}{5pt}
    \begin{tabular}{l l *{4}{r}}
    \toprule
        $\mathcal{T}_{\text{train}}$ & $\mathcal{T}_{\text{test}}$ & \textbf{Adapter} & \textbf{LoRA} & \textbf{Prefix} & \textbf{Avg.} \\
        \midrule
        \multirow{2}{*}{SST-2} & Rotten Tomatoes & 99.3 & 98.5 & 95.5 & 97.8 \\
         & Amazon Review & 96.7 & 96.5 & 97.1 & 96.8 \\
        \midrule
        \multirow{2}{*}{MNLI} & SciTail & 82.0 & 74.1 & 73.3 & 76.5 \\
        & RTE & 71.5 & 80.0 & 71.3 & 74.3 \\
        \midrule
        \multirow{1}{*}{WiC} & WSC & 61.8 & 84.8 & 85.7 & 77.4 \\
        \midrule
        \multirow{1}{*}{QQP} & MRPC & 85.4 & 81.5 & 68.5 & 78.5 \\
        \midrule
        \multirow{2}{*}{ELI5-ELI5} & ELI5-Askh & 89.1 & 85.1 & 77.5 & 83.9 \\
        & ELI5-Asks & 93.9 & 94.4 & 88.2 & 92.2 \\
        \midrule
        \multirow{3}{*}{DREAM} & CODAH & 63.6 & 63.9 & 61.6 & 63.0 \\
        & QuaRTz & 74.2 & 72.4 & 73.9 & 73.5 \\
        & CoPA & 90.4 & 66.0 & 101.4 & 85.9 \\
        \midrule
        & \textbf{Avg.} & 82.5 & 81.6 & 81.3 & 81.8 \\
        \bottomrule
    \end{tabular}
    \caption{Relative performance (\%) for subspace optimization under single-task setting with constructed subspace.}
    \label{tab:share-intrinsic-single-task-subspace-optimization}
\end{table}
\begin{table}[t!]
    \centering
    \scriptsize
    \begin{tabular}{l *{4}{r}}
    \toprule
        $\mathcal{T}_{\text{test}}$ & \textbf{Adapter} & \textbf{LoRA} & \textbf{Prefix} & \textbf{Avg.} \\
        \midrule
        Rotten Tomatoes & 92.9 & 88.5 & 66.4 & 82.6\\
        Yelp Polarity & 97.0 & 96.2 & 90.0 & 94.4\\
        \midrule
        WSC & 94.1 & 90.9 & 82.9 & 89.3\\
        \midrule
        AI2 ARC & 77.2 & 73.4 & 75.9 & 75.5\\
        QASC & 73.8 & 64.0 & 59.8 & 65.9\\
        QuaRTz & 78.0 & 78.5 & 74.3 & 76.9\\
        \midrule
        BLiMP-ANA & 94.0 & 98.0 & 47.0 & 79.7\\
        \midrule
        ELI5-Asks & 83.9 & 84.6 & 79.5 & 82.7\\
        \midrule
        ETHOS-Gender & 54.6 & 55.4 & 74.3 & 61.4\\
        \midrule
        \textbf{Avg.} & 82.8 & 81.1 & 72.2 & 78.7\\
        \bottomrule
    \end{tabular}
    \caption{Relative performance (\%) for subspace solution transfer under under multi-task setting with constructed subspace.}
    \label{tab:share-intrinsic-multi-task-subspace-optimization}
\end{table}

From a standpoint of analysis, it is necessary to start our pipeline from independent solutions of different \PETs and then explore their connections. Now that we have validated the existence of the unified optimization subspace, for practical uses, we could simplify the original pipeline by enforcing different \PETs to share the same intrinsic vector. Specifically, we \textit{jointly} train three \PETs, and generate the parameters of each \PET via a \textit{shared} intrinsic vector and three individual up-projections. Both the intrinsic vector and the up-projections are trainable. Denote the intrinsic vector shared among \PETs on the $i$-th task as $\mathbf{I}^i_\text{shared}$, then the parameters of \PET $t_*$ for the $i$-th task are generated as $\mathcal{\theta}_{t_*}^i=\texttt{Proj}_{t_*}^{\uparrow}(\mathbf{I}^i_\text{shared})$. During the joint training, we minimize the loss:
\begin{align*}
\mathcal{L}&=\frac{1}{3}\sum_{i=1}^{|\mathcal{T}_{\text{train}}|} \sum_{t_*\in\{t_\text{A}, t_\text{P}, t_\text{L}\}}\mathcal{L}_{\text{task}}^i(\overline{\theta_{t_*}^i}\mid\theta_0).
\end{align*}
In this way, we can directly approximate the desired unified subspace, omitting the procedure of first obtaining solutions for different \PETs, and we do not need to assign a down-projection for each \PET. Subspace optimization and subspace solution transfer can then be carried out using this subspace.

\begin{table}[t!]
    \centering
    \scriptsize
    \setlength{\tabcolsep}{1pt}
    \begin{tabular}{l l *{7}{r}}
    \toprule
        $\mathcal{T}_{\text{train}}$ & $\mathcal{T}_{\text{test}}$ & $\mathbf{A} \!\!\rightarrow\!\! \mathbf{L}$ & $\mathbf{A} \!\!\rightarrow\!\! \mathbf{P}$ & $\mathbf{L} \!\!\rightarrow\!\! \mathbf{A}$ & $\mathbf{L} \!\!\rightarrow\!\! \mathbf{P}$ & 
        $\mathbf{P} \!\!\rightarrow\!\! \mathbf{A}$ & 
        $\mathbf{P} \!\!\rightarrow\!\! \mathbf{L}$ & \textbf{Avg.} \\
        \midrule
        \multirow{2}{*}{SST-2} & R. Tomatoes & 99.7 & 98.5 & 100.9 & 98.8 & 101.2 & 99.7 & 99.8 \\
        & A. Review & 98.2 & 98.2 & 98.3 & 98.2 & 98.1 & 97.3 & 98.1 \\
        \midrule
        \multirow{2}{*}{MNLI} & SciTail & 74.9 & 77.4 & 83.5 & 77.0 & 83.9 & 74.8 & 78.6 \\
        & RTE & 89.1 & 80.2 & 58.7 & 87.1 & 70.6 & 87.3 & 78.8 \\
        \midrule
        \multirow{1}{*}{WiC} & WSC & 84.8 & 74.3 & 70.7 & 74.3 & 82.3 & 72.8 & 76.5 \\
        \midrule
        \multirow{1}{*}{QQP} & MRPC & 84.8 & 67.4 & 79.8 & 76.4 & 78.0 & 5.6 & 65.3 \\
        \midrule
        \multirow{2}{*}{ELI5-ELI5} & ELI5-Askh & 85.1 & 79.9 & 93.4 & 79.9 & 92.6 & 85.9 & 86.1 \\
        & ELI5-Asks & 93.1 & 91.5 & 97.9 & 91.5 & 97.3 & 93.8 & 94.2 \\
        \midrule
        \multirow{3}{*}{DREAM} & QuaRTz & 80.0 & 74.9 & 77.7 & 74.2 & 78.1 & 80.2 & 77.5 \\
        & CODAH & 73.0 & 77.8 & 75.9 & 78.2 & 80.3 & 70.9 & 76.0 \\
        & CoPA & 67.7 & 107.5 & 86.8 & 103.8 & 84.4 & 63.8 & 85.7 \\
        \midrule
        & \textbf{Avg.} & 84.6 & 84.3 & 84.0 & 85.4 & 86.1 & 75.6 & 83.3 \\
        \bottomrule
    \end{tabular}
    \caption{Relative performance (\%) for subspace solution transfer under single-task setting with constructed subspace.}
    \label{tab:share-intrinsic-single-task-solution-transfer}
\end{table}
\begin{table}[t!]
    \centering
    \scriptsize
    \setlength{\tabcolsep}{2pt}
    \begin{tabular}{l *{10}{r}}
    \toprule
          $\mathcal{T}_{\text{test}}$ &
          $\mathbf{A} \!\!\rightarrow\!\! \mathbf{L}$ & $\mathbf{A} \!\!\rightarrow\!\! \mathbf{P}$ & $\mathbf{L} \!\!\rightarrow\!\! \mathbf{A}$ & $\mathbf{L} \!\!\rightarrow\!\! \mathbf{P}$ & 
        $\mathbf{P} \!\!\rightarrow\!\! \mathbf{A}$ & 
        $\mathbf{P} \!\!\rightarrow\!\! \mathbf{L}$ & \textbf{Avg.} \\ 
        \midrule
        Rotten Tomatoes& 93.6 & 88.0 & 83.5 & 87.9 & 84.6 & 95.0 & 88.8\\
        Yelp Polarity & 97.5 & 92.0 & 98.0 & 87.0 & 97.9 & 97.1 & 94.9\\
        \midrule
        WSC & 93.9 & 94.3 & 58.9 & 94.3 & 64.7 & 93.9 & 83.3\\
        \midrule
        AI2 ARC & 82.7 & 78.3 & 88.4 & 77.7 & 83.6 & 81.4 & 82.0\\
        QASC & 85.2 & 68.2 & 103.2 & 70.7 & 99.9 & 85.2 & 85.4\\
        QuaRTz & 79.1 & 76.9 & 81.4 & 76.4 & 81.4 & 77.4 & 78.8\\
        \midrule
        BLiMP-ANA & 99.0 & 51.0 & 96.0 & 52.0 & 96.0 & 96.0 & 81.7\\
        \midrule
        ELI5-Asks & 90.5 & 89.5 & 89.9 & 88.8 & 87.9 & 90.5 & 89.5\\
        \midrule
        ETHOS-Gender & 52.8 & 0.0 & 51.1 & 0.0 & 50.1 & 57.4 & 35.2\\
        \midrule
        \textbf{Avg.} & 86.0 & 70.9 & 83.4 & 70.5 & 82.9 & 86.0 & 80.0 \\
        \bottomrule
    \end{tabular}
    \caption{Relative performance (\%) for subspace solution transfer under multi-task setting with constructed subspace.}
    \label{tab:share-intrinsic-multi-task-solution-transfer}
\end{table}

We present the results of subspace optimization and subspace solution transfer for both the single-task and multi-task settings in~\cref{tab:share-intrinsic-single-task-subspace-optimization,tab:share-intrinsic-single-task-solution-transfer,tab:share-intrinsic-multi-task-subspace-optimization,tab:share-intrinsic-multi-task-solution-transfer}. We find that in general, all \PETs still achieve non-trivial performance in both subspace optimization and subspace solution transfer, which means the simplification does not influence the representation ability of the found subspace. This simplified pipeline is the cornerstone of our analysis of the connection between fine-tuning and \PETs. Since the simplified procedure does not require training a down-projection, the total number of tunable parameters can be further reduced.
% transferability of \PETs generally becomes poorer in both single-task and multi-task settings. This may be because sharing the intrinsic vector limits the capability of the \PETs.

\subsection{HyperParameters of simplified approximation experiments}
\label{sec:appendix-results-share-intrinsic}

In the simplified approximation experiments, we use different learning rates for the shared intrinsic vector and \PETs in subspace approximation. We set the learning rate as $5 \times 10^{-5}$ for the shared intrinsic vector, and $1 \times 10^{-4}$ for \PETs. The batch size is $16$ in the single-task setting and $8$ in the multi-task setting. We train the model for a maximum of $100000$ and validate  every $1000$ steps. For subspace optimization, we perform grid search on learning rate in \{$5 \times 10^{-2}$, $1 \times 10^{-2}$, $5 \times 10^{-3}$, $1 \times 10^{-3}$ \}. We set batch size as $16$.
% use the up-projections from the best checkpoint of subspace approximation, and we 
To keep 3 \PETs' number of parameters consistent, we set $r_\text{A}$ as $12$, $r_\text{L}$ as $10$, $m$ as $24$, and $d_\text{P}$ as $120$. Other hyperparameters are kept consistent with the main experiments.
% The results for subspace optimization under single-task setting is listed in~\cref{tab:share-intrinsic-single-task-subspace-optimization}. 

\begin{table}[t!]
    \centering
    \scriptsize
    \setlength{\tabcolsep}{5pt}
    \begin{tabular}{l@{~~~}l@{~~~}r@{~~~}r@{~~~}r@{~~~}r}
    \toprule
        $\mathcal{T}_{\text{train}}$ & $\mathcal{T}_{\text{test}}$ & \textbf{Adapter} & \textbf{LoRA} & \textbf{Prefix} & \textbf{Fine-tune} \\
        \midrule
        \multirow{2}{*}{SST-2} & Rotten Tomatoes & $89.2$ & $89.3$ & $90.0$ & $89.8$ \\
         & Amazon Review & $96.2$ & $96.6$ & $96.6$ & $97.0$ \\
        \midrule
        \multirow{2}{*}{MNLI} & SciTail & $94.0$ & $93.8$ & $93.0$ & $94.8$ \\
        & RTE & $78.4$ & $79.1$ & $72.7$ & $80.6$ \\
        \midrule
        \multirow{1}{*}{WiC} & WSC & $65.4$ & $63.5$ & $67.3$ & $67.3$ \\
        \midrule
        \multirow{1}{*}{QQP} & MRPC & $87.3$ & $87.3$ & $87.3$ & $89.7$ \\
        \midrule
        \multirow{2}{*}{ELI5-ELI5} & ELI5-Askh & $11.5$ & $12.0$ & $12.6$ & $13.0$ \\
        & ELI5-Asks & $15.0$ & $15.2$ & $15.1$ & $15.3$ \\
        \midrule
        \multirow{3}{*}{DREAM} & CODAH & $41.2$ & $43.0$ & $45.0$ & $45.2$ \\
        & QuaRTz & $67.0$ & $67.1$ & $68.5$ & $69.4$ \\
        & CoPA & $60.4$ & $56.4$ & $58.4$ & $59.2$ \\
        \midrule
        & \textbf{Avg.} & $65.6$ & $65.2$ & $65.0$ & $66.6$\\
        \bottomrule
    \end{tabular}
    \caption{Absolute performance for different tuning methods under the single-task setting.}
    \label{tab:single-task-baseline}
\end{table}

\subsection{Implementation Details for Extension to Fine-tuning}
\label{sec:detail_fastfood}
We use the simplified pipeline introduced in~\cref{sec:appendix-shared-intrinsic} to further reduce the number of trainable parameters. That is, an intrinsic vector $\mathbf{I}^i$ for the $i$-th task is set to be a trainable parameter, and is shared among fine-tuning and different \PETs. The steps of analysis is the same as in~\cref{sec:experiment-results}.
In the experiment of subspace approximation on glue-sst2. We set learning rate as 1e-4, batch size as 8, max steps as 100000 and validate every 1000 steps. For subspace optimization, we perform grid search on learning rate in \{$1 \times 10^{-1}$, $5 \times 10^{-2}$, $1 \times 10^{-2}$, $5 \times 10^{-3}$, $1 \times 10^{-3}$, $5 \times 10^{-4}$, $1 \times 10^{-4}$ \}. We set batch size as 8 and validate every 100 steps. Other hyper-parameters are the same as ~\cref{sec:appendix-results-share-intrinsic}.

\subsection{Absolute Performance for Different Tuning Methods}
In the main paper, we report the relative performance of subspace optimization. In this section, we list the absolute performance of different tuning methods (Adapter, Prefix-Tuning, and LoRA) in Table~\ref{tab:single-task-baseline} and Table~\ref{tab:multi-task-baseline} for reference.

\begin{table}[t!]
    \centering
    \scriptsize
    \begin{tabular}{l *{4}{r}}
    \toprule
        $\mathcal{T}_{\text{test}}$ & \textbf{Adapter} & \textbf{LoRA} & \textbf{Prefix} & \textbf{Fine-tune} \\
        \midrule
        Rotten Tomatoes & $89.2$ & $89.3$ & $90.0$ & $89.8$\\
        Yelp Polarity & $97.3$ & $97.4$ & $97.8$ & $97.9$\\
        \midrule
        WSC & $65.4$ & $63.5$ & $67.3$ & $67.3$\\
        \midrule
        AI2 ARC & $31.2$ & $32.4$ & $32.2$ & $31.3$\\
        QASC & $33.0$ & $37.8$ & $33.3$ & $43.6$\\
        QuaRTz & $67.0$ & $67.1$ & $68.5$ & $69.4$\\
        \midrule
        BLiMP-ANA & $100.0$ & $100.0$ & $100.0$ & $100.0$\\
        \midrule
        ELI5-Asks & $15.0$ & $15.2$ & $15.1$ & $15.3$\\
        \midrule
        ETHOS-Gender & $79.9$ & $79.9$ & $77.4$ & $74.5$\\
        \midrule
        \textbf{Avg.} & $65.6$ & $66.2$ & $65.4$ & $67.0$\\
        \bottomrule
    \end{tabular}
    \caption{Absolute performance for different tuning methods under the multi-task setting.}
    \label{tab:multi-task-baseline}
\end{table}

\begin{table*}[!ht]
  \centering
  \small
  \caption{The training tasks involved in our multi-task setting.}
    \begin{tabular}{clll}
    \toprule
    \textbf{Split} & \textbf{Task Name} & \textbf{Reference}\\
    \midrule
    \multirow{60}[2]{*}{Training tasks}
          & amazon review & \citealt{mcauley2013hidden} \\
          & financial\_phrasebank & \citealt{malo2014good} \\
          & glue-sst2 & \citealt{socher-etal-2013-recursive} \\
          & imdb  & \citealt{maas-etal-2011-learning}   \\
          & emotion &	\citealt{saravia-etal-2018-carer} \\
          & tweet\_eval-offensive &	\citealt{barbieri-etal-2020-tweeteval} \\
          & tweet\_eval-stance\_climate &	\citealt{barbieri-etal-2020-tweeteval} \\
          &   ethos-directed\_vs\_generalized   &	\citealt{Mollas2020ETHOSAO} \\
          &   ethos-race   &	\citealt{Mollas2020ETHOSAO} \\
          &   hatexplain &	\citealt{mathew2020hatexplain}  \\
          &  glue-mnli & \citealt{williams-etal-2018-broad}   \\
          &  glue-qnli &	\citealt{rajpurkar-etal-2016-squad}   \\
          &  glue-wnli &	\citealt{faruqui-das-2018-identifying}   \\
          &  superglue-rte & \begin{tabular}[c]{@{}l@{}}\citealt{dagan2005pascal, bar2006second}\\\citealt{giampiccolo2007third, bentivogli2009fifth}\end{tabular}   \\
          &   health\_fact &	\citealt{kotonya-toni-2020-explainable-automated}  \\
          &   liar  &	\citealt{wang-2017-liar} \\
          &   glue-qqp & \href{https://quoradata.quora.com/First-Quora-Dataset-Release-Question-Pairs}{(link)}  \\
          &  medical\_questions\_pairs & \citealt{medical-qqp}  \\
          &  paws & \citealt{zhang-etal-2019-paws}   \\
          &  circa  &	\citealt{louis2020d}   \\
          &  onestop\_english  &	\citealt{vajjala2018onestopenglish}   \\
          &  trec-finegrained  &	\citealt{li-roth-2002-learning}; \citealt{hovy-etal-2001-toward}   \\
          &  wiki\_auto  &	\citealt{jiang-etal-2020-neural}   \\
          &   google\_wellformed\_query &	\citealt{faruqui-das-2018-identifying}  \\
          &   sms\_spam &	\citealt{sms_spam}   \\
          &   superglue-wic  &	\citealt{pilehvar-camacho-collados-2019-wic} \\
          &  lama-google\_re &	\citealt{petroni-etal-2019-language,petroni2020how}   \\
          &  numer\_sense &	\citealt{lin-etal-2020-birds}   \\
          &  search\_qa  &	\citealt{Dunn2017SearchQAAN}  \\
          &  web\_questions  &	\citealt{berant-etal-2013-semantic}  \\
          &  boolq &	\citealt{clark-etal-2019-boolq}   \\
          &  codah  &	\citealt{chen-etal-2019-codah}  \\
          &  commonsense\_qa &	\citealt{talmor-etal-2019-commonsenseqa}   \\
          &  cosmos\_qa  &	\citealt{huang-etal-2019-cosmos}  \\
          &  dream &	\citealt{saha-etal-2018-duorc}   \\
          &  hellaswag &	\citealt{zellers-etal-2019-hellaswag}   \\
          &  sciq  &	\citealt{welbl-etal-2017-crowdsourcing}  \\
          &  quail  &	\citealt{Rogers_Kovaleva_Downey_Rumshisky_2020}  \\
          &  quarel &	\citealt{Tafjord_Clark_Gardner_Yih_Sabharwal_2019}   \\
          &  race-high &	\citealt{lai-etal-2017-race}   \\
          &  superglue-copa  &	\citealt{gordon-etal-2012-semeval}  \\
          &  wino\_grande &	\citealt{Sakaguchi_Le_Bras_Bhagavatula_Choi_2020}   \\
          &  eli5-eli5  & \citealt{fan-etal-2019-eli5}  \\
          &  hotpot\_qa  &	\citealt{yang-etal-2018-hotpotqa}  \\
          &  quoref &	\citealt{dasigi-etal-2019-quoref}   \\
          &  superglue-record  & \citealt{Zhang2018ReCoRDBT}  \\
          &  multi\_news & \citealt{fabbri-etal-2019-multi}   \\
          &  xsum &	\citealt{narayan-etal-2018-dont}   \\
          &  spider &	\citealt{yu-etal-2018-spider}   \\
          &  wikisql &	\citealt{zhongSeq2SQL2017}   \\
          & blimp-anaphor\_gender\_agreement  & \citealt{warstadt2019blimp}  \\
          & blimp-ellipsis\_n\_bar\_1 & \citealt{warstadt2019blimp}   \\
          & blimp-irregular\_past\_participle\_adjectives & \citealt{warstadt2019blimp}   \\
          & blimp-wh\_questions\_object\_gap & \citealt{warstadt2019blimp}   \\
          & cos\_e & \citealt{rajani-etal-2019-explain}  \\
          &   acronym\_identification  & \citealt{pouran-ben-veyseh-etal-2020-acronym}  \\
          &  crawl\_domain &	\citealt{zhang-etal-2020-semi}   \\
          &  proto\_qa &	\citealt{boratko-etal-2020-protoqa}   \\
          &  qa\_srl &	\citealt{he-etal-2015-question}   \\
    \bottomrule
    \end{tabular}%
  \label{tab:multitask_train}%
\vspace{-2em}
\end{table*}%

\begin{table*}[!ht]
  \centering
  \small
  \caption{The test tasks involved in our multi-task setting.}
\begin{tabular}{clll}
    \toprule
    \textbf{Split} & \textbf{Task Name} & \textbf{Reference}\\
    \midrule
    \multirow{9}[2]{*}{Test tasks} 
          & rotten\_tomatoes & \citealt{pang-lee-2005-seeing} \\
          & yelp\_polarity & \citealt{zhang2015character} \\
          &   ethos-gender   &	\citealt{Mollas2020ETHOSAO} \\
          &   superglue-wsc &	\citealt{levesque2012winograd}  \\
          &  ai2\_arc &	\citealt{Clark2018ThinkYH}   \\
          &  qasc  &	\citealt{khot2020qasc}  \\
          &  quartz-no\_knowledge &	\citealt{tafjord-etal-2019-quartz}   \\
          &  eli5-asks & \citealt{fan-etal-2019-eli5}   \\
          & blimp-anaphor\_number\_agreement
          & \citealt{warstadt2019blimp}   \\
        \bottomrule
    \end{tabular}%
  \label{tab:multitask_test}%
\vspace{-2em}
\end{table*}%

\end{document}